\documentclass[runningheads]{llncs}
\usepackage{graphicx}

\usepackage{enumitem}
\usepackage{amsmath,amssymb,amsfonts}
\usepackage{multirow}
\usepackage{booktabs}
\usepackage{algorithmic}
\usepackage[ruled,vlined]{algorithm2e}
\usepackage{color}
\usepackage{wrapfig}
\usepackage[misc]{ifsym} 

\usepackage{enumitem}
\setlist{nolistsep}

\makeatletter
\g@addto@macro\normalsize{%
\setlength\abovedisplayskip{3pt}
\setlength\belowdisplayskip{2pt}
\setlength\abovedisplayshortskip{3pt}
\setlength\belowdisplayshortskip{2pt}
}
\makeatother

\begin{document}
\title{Model Inversion Attacks on Homogeneous and Heterogeneous Graph Neural Networks\thanks{This paper was accepted by SecureComm 2023.}}
\titlerunning{Model Inversion Attacks on Homo/Hetero-geneous Graph Neural Networks}
%

\author{Renyang Liu\inst{1}\orcidID{0000-0002-7121-1257} \and
Wei Zhou\inst{1}\orcidID{0000-0002-5881-9436} \and
Jinhong Zhang\inst{1}\orcidID{0000-0002-9906-3508} \and
Xiaoyuan Liu\inst{2}\orcidID{0000-0002-2625-3896} \and
Peiyuan Si\inst{3}\orcidID{0000-0002-0739-0107} \and
Haoran Li\inst{\textsuperscript{1,~\Letter}}  \orcidID{0000-0002-0409-2227}}
\authorrunning{Renyang Liu et al.}

\institute{Yunnan University, Kunming, Yunnan, China\\
\email{\{ryliu,jhnova,lihaoran\}@mail.ynu.edu.cn,zwei@ynu.edu.cn} \\ 
\and
University of Electronic Science and Technology of China, Chengdu, Sichuan, China\\
\email{xiaoyuanliu@std.uestc.edu.cn} \\
\and
Nanyang Technological University, Singapore \\
\email{peiyuan001@ntu.edu.sg}}

\maketitle              

\begin{abstract}
Recently, Graph Neural Networks (GNNs), including Homogeneous Graph Neural Networks (HomoGNNs) and Heterogeneous Graph Neural Networks (HeteGNNs), have made remarkable progress in many physical scenarios, especially in communication applications. Despite achieving great success, the privacy issue of such models has also received considerable attention. Previous studies have shown that given a well-fitted target GNN, the attacker can reconstruct the sensitive training graph of this model via model inversion attacks, leading to significant privacy worries for the AI service provider. We advocate that the vulnerability comes from the target GNN itself and the prior knowledge about the shared properties in real-world graphs. Inspired by this, we propose a novel model inversion attack method on HomoGNNs and HeteGNNs, namely HomoGMI and HeteGMI. Specifically, HomoGMI and HeteGMI are gradient-descent-based optimization methods that aim to maximize the cross-entropy loss on the target GNN and the $1^{st}$ and $2^{nd}$-order proximities on the reconstructed graph. Notably, to the best of our knowledge, HeteGMI is the first attempt to perform model inversion attacks on HeteGNNs. Extensive experiments on multiple benchmarks demonstrate that the proposed method can achieve better performance than the competitors.
\keywords{Model Inversion Attack \and Adversarial Attack \and  Graph Neural Network \and Graph Representation Learning \and Network Communication.}
\end{abstract}

\section{Introduction}\label{Introduction}
Graph Neural Networks (GNNs) have various applications in communication systems, including Network Traffic Analysis, Resource Allocation, Wireless Channel Modeling, Network Routing and Optimization, and Network Representation Learning \cite{DBLP:journals/comcom/Jiang22,DBLP:journals/tmc/LeeYD23}. Besides, many real-world data in our daily life can be modeled as graphs, which are mathematical structures consisting of nodes and edges that connect the nodes. According to the assumptions on graph structures, these graphs can be divided into homogeneous graphs (HomoGs) and heterogeneous graphs (HeteGs), where HeteGs contain multiple types of nodes and edges. Both of HomoGs and HeteGs are commonly used in social networks, citation networks and biological networks. Starting from the analysis requirement of such data, GNNs including Homogeneous Graph Neural Networks (HomoGNNs) and Heterogeneous Graph Neural Networks (HeteGNNs), have achieved extensive research attention and also shown outstanding performance in various applications, including recommended systems malicious detection \cite{hou2017hindroid,ijcai2019-522}, event prediction \cite{luo2020dynamic,zheng2020heterogeneous}, financial services\cite{xiang2022temporal,zheng2021heterogeneous}, etc. 

\begin{figure}[ht]
    \centering
    \includegraphics[width=0.6\textwidth]{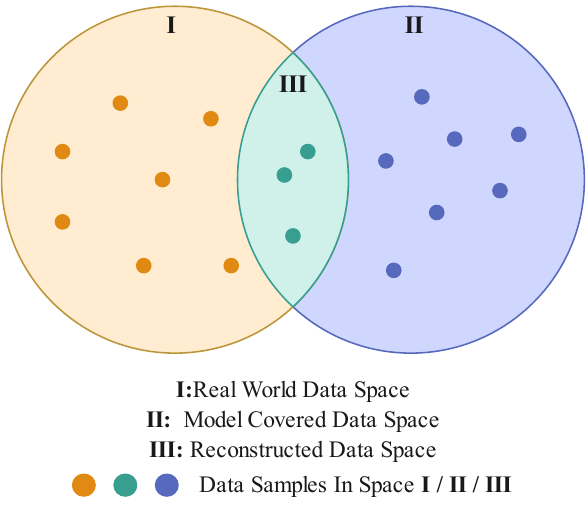}
    \caption{We believe that the reconstructed data should match the characteristics of a real-world graph since it does not make sense to use a fantasy graph to train a model. This goal is achieved by making the reconstructed graph satisfy first-order and second-order proximities. We also believe that the reconstructed data should fall within the model-covered space, i.e., when we feed the reconstructed data into the target model, the model should return a lower loss value.}
    \label{MIA}
\end{figure}

Since GNNs have been deployed in various real-world applications \cite{fan2019metapath,liu2020heterogeneous,hu2018leveraging}, the privacy and security issues of such models have received considerable attention. GNNs are proved to be vulnerable to model inversion attacks \cite{graphmiTrans,graphmi,DBLP:journals/popets/OlatunjiRFK23}, that is, given a well-trained GNN model, the attacker can reconstruct the entire private training graph by exploiting the auxiliary knowledge and the outputs of this GNN for some unlawful purpose. Such vulnerability of GNNs may cause tremendous security risks, especially in the privacy-critical area. Take credit evaluation as an example, attacker may build the private transaction records, so as to extortion banks and cause serious economic losses.

Studying model inversion attack on GNNs helps us understand the potential vulnerability of GNNs and enable us to evaluate and avoid privacy risks in real-world applications. Thus, in this paper, we aim to design a novel model inversion attack method applicable to both HomoGNNs and HeteGNNs. To guide the design of the attack method, we first need to investigate a critical question: \textit{What our reconstructed data should look like?} A straightforward answer is that the data reconstructed by the model inversion attack should be similar to the training data used by this model, i.e., they have similar data space. However, there is no silver bullet here to directly measure the margin between the space of the reconstructed data and the training data, let alone that our goal is to reconstruct the training data itself. This instinct reminds us that we need to rely on some prior knowledge to infer the space of the training data in a non-directive way.

There is a consensus that the training data we use to train our model is derived from the real world but can only cover part of the space of real-world data. For a simple illustration, we label the real-world data space as \textbf{I} (shown in Fig. \ref{MIA}). After training, the model will be well-fitted with the training data. However, this well-trained model may also cover some data out of the real-world space, e.g., adversarial examples, which are designed artificially, but the model still treats them accordingly as real-world data. In fact, some works indicted that adversarial examples are ubiquitous because the neural networks' vulnerabilities cannot be eradicated \cite{DBLP:journals/corr/abs-2203-17209,DBLP:journals/tnn/YuanHZL19}. We mark the space that the model can cover as \textbf{II} (Fig. \ref{MIA}).

The above analyses inform us that the space of the data samples reconstructed by the model inversion attack should fall at the intersection of the space \textbf{I} and the space \textbf{II}; we mark this intersection as space \textbf{III} (Please refer to Fig. \ref{MIA}). Hence, the following issue is how to propel the reconstructed data to fall into space \textbf{I} and \textbf{II}. 

In this paper, we treat the model inversion attack on graph-structured data as an optimization problem and propose two components to make the reconstructed data in space \textbf{I} and space \textbf{II}. Specifically, in order to make the reconstructed graph fall into space \textbf{I}, we devote to making the reconstructed graph have a higher accuracy when it has been fed into the target GNN model. Besides, we consider maximizing $1^{st}$ and $2^{nd}$ order proximities of the reconstructed graph to make it have shared characteristics with the real-world graphs and hence make the reconstructed data fall into space \textbf{II}. In various graph embedding methods \cite{tang2015line,ribeiro2017struc2vec}, $1^{st}$ and $2^{nd}$ order proximities often appear as part of the objective function, be used to make the embedding result preserve the topological information on the graph better. To some extent, the success of these graph embedding methods can corroborate the widespread existence of  $1^{st}$ and $2^{nd}$ order proximities in real-world graphs.

Based on the above ideas, two model inversion attack methods, called Homogeneous Graph Model Inversion Attack (HomoGMI) and Heterogeneous Graph Model Inversion Attack (HeteGMI), have been developed to perform the model inversion attacks on HomoGNNs and HeteGNNs. To summarize, in this paper, our contributions can be written as follows: 
\begin{itemize}[leftmargin=*]
    \item We propose a hypothesis in which the reconstructed data should fall into both the real-world data space and the model covered data space. Based on this assumption, we proposed HomoGMI. Different from previous works, we first introduced  $1^{st}$ and $2^{nd}$ order proximities into model inversion attacks against GNNs.
    \item Based on HomoGMI, a novel model inversion attack method applicable to HeteGNNs, called HeteGMI, is proposed to reconstruct the private training graph. To the best of our knowledge, HeteGMI is the first model inversion attack method on HeteGNNs.
    \item Extensive experimental results on various public datasets have demonstrated the effectiveness of HomoGMI and HeteGMI concerning various metrics. Notably, we find that the existing model inversion methods designed for HomoGNNs were not effective when applied to the HeteGNNs, while HeteGMI is the only valid model inversion attack method.
\end{itemize}

The rest of this paper is organized as follows. In Sec. \ref{sec:background}, we briefly introduce the basic concepts of HomoG and HeteG (and their corresponding GNNs) and discuss existing model inversion attack methods on GNN models of graph-structured data. The techniques of the proposed attack methods on HomoGNNs and HeteGNNs are given in Sec. \ref{sec:homoGNN} and Sec. \ref{sec:heteGNN}. In Sec. \ref{sec:experiments}, we discuss the experiments, including the basic settings, the compared methods and the experimental results. Sec. \ref{sec:conclusion} draws the conclusions of this paper.

\section{Background}
\label{sec:background}
\subsection{Definitions and Notations}
\label{Definitions and Notations}
In this section, we briefly describe the concepts, definitions and related variables involved in this paper.

\textbf{Homogeneous Graph (HomoG)}:
An undirected and unweighted graph is denoted as $\mathcal{G} = \{\mathcal{V}, \mathcal{E}\}$, where $\mathcal{V}$ and $\mathcal{E}$ denotes the node set and the edge set of the graph. $A$ is a symmetric adjacency matrix $A=\{0,1\}^{|\mathcal{V}|\times|\mathcal{V}|}$, and $A_{ij}$ is non-zero if there exists a edge between node $v_i$ and node $v_j$. There may exist $d$-dimensional features for each node in this graph, and these features can be described by a feature matrix $X \in \mathbb{R}^{|\mathcal{\mathcal{V}}| \times d}$. In this scenario, a graph can also be denoted as $\mathcal{G} = (A, X)$.\\

\textbf{Heterogeneous Graph (HeteG)}:
Compared with HomoG, the main difference of HeteG is that each node $v \in \mathcal{\mathcal{V}}$ and each edge $e \in \mathcal{E}$  in a HeteG are associated with their mapper functions $\phi(v): \mathcal{V} \rightarrow \mathcal{T}$ and $\psi(e): \mathcal{E} \rightarrow \mathcal{C}$, where $\mathcal{T}$ and $\mathcal{C}$ denote the node types and edge types, and $|\mathcal{T}| + |\mathcal{C}| > 2$. \\

\textbf{Edge-type Specified Matrix}: The edge-type specified matrix is build from the sub-graph of a HeteG by specifying the relations to retain. Given a edge type $c$, the edge-type specified matrix is denoted by $A^c$. Note that we can build the full adjacency matrix $A$ from ${A^{c_1}, A^{c_2}, \dots, A^{c_K}}$ easily. \\

\textbf{Meta-path}:
A meta-path $m$ on HeteG is defined as a finite sequence of the node type, that is $m = t_1 \xrightarrow{c_1} t_2 \xrightarrow{c_2} ... \xrightarrow{c_K} t_{K+1}$ (simplified to $m = t_1 t_2 \dots t_{K+1}$), where the node types $t_1, t_2, ..., t_{K+1} \in \mathcal{T}$ and the edge types $c_1, c_2, ..., c_{K} \in \mathcal{C}$. \\

\textbf{Meta-path Augmented Adjacency Matrix}:
Given a meta-path $m$, we define a meta-path augmented adjacency matrix as $W^m = [w^m_{ij}]$, where $w^m_{ij}$ is the number of path following $m$ connecting nodes $v_i$ and $v_j$. If $m$ is a symmetric meta-path, the corresponding $W^m$ is a symmetric matrix also, and in this scenario, a homogeneous sub-graph can be generated from the $W^m$.
\begin{figure}
    \centering
    \includegraphics[width=0.6\textwidth]{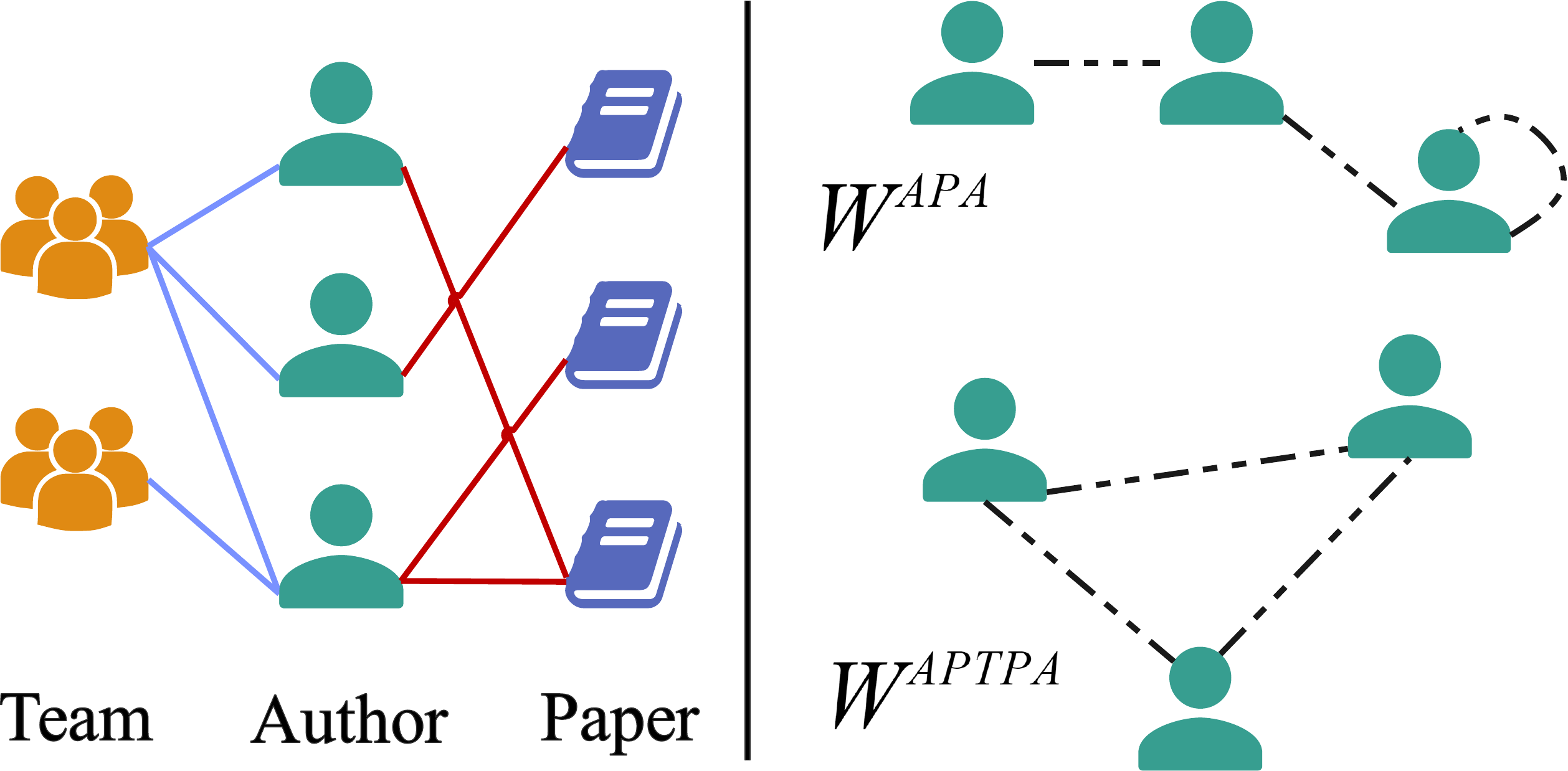} 
    \caption{A example of an undirected HeteG (left), which contains three node types (Team (T), Author (A) and Paper (P)) and two edge types (TA, AP). Given a set of meta-path, say, APA and APTPA, the $W^{APA}$ and $W^{APTPA}$ indicated graph has shown on the right.}
    \label{graph_example_mi}
\end{figure}

For example, in the HeteG that shown in Fig. \ref{graph_example_mi}, given a symmetric meta-path $m$, say, $m = APA$, then the $W^{APA}$ can be obtained as follows:
\begin{equation}
\label{computew}
    \begin{aligned}
            W^{APA} &= A^{AP}A^{PA}, \\
            \text{that is:}\ \ \ \ \ \ \ \ \ \ \ \ & \\
            \left[ \begin{array}{ccc}
        1 & 1 & 1\\
        1 & 2 & 1\\
        1 & 1 & 2
        \end{array} 
        \right ] &= \left[ \begin{array}{ccc}
        0 & 1 & 0\\
        1 & 1 & 0\\
        0 & 1 & 1
        \end{array} 
        \right ] \left[ \begin{array}{ccc}
        0 & 1 & 0\\
        1 & 1 & 1\\
        0 & 0 & 1
        \end{array} 
        \right ].
    \end{aligned}
\end{equation}
where $A^{PA} = (A^{AP})^T$ is the edge-type specified matrix. Note that $W^{AP}$ and $W^{PA}$ may not be a square matrix.

\subsection{$1^{st}$ and $2^{nd}$-order Proximities on Graphs}
Many prior works have demonstrated that real-world graphs have the same shared properties, which the properties can be modeled as the $1^{st}$ order proximity and the $2^{nd}$ order proximity. 

The meaning of the $1^{st}$ order proximity on graph data is that two nodes are considered similar if they are connected by an edge, which the proximity is widely available in the real world, for example, people who make friends with each other in social networks tend to have similar interests; pages linked to each other on the world wide web tend to discuss similar topics.

The $2^{nd}$ order proximity on graph data says two nodes are considered similar if they share many neighbouring nodes, which is also widely available in the real world. For example, in social networks, people who share similar friends tend to have similar interests and thus become friends; in word co-occurrence networks, words that always co-occur with the same set of words tend to have similar meanings.

In graph embedding methods, $1^{st}$ and $2^{nd}$-order proximities are widely adopted \cite{s2v,line,MetaPathBasedProximity}, these methods usually attempt to make two similar nodes indicated by the $1^{st}$ and $2^{nd}$-order proximities, close to each other in the embedding space.

\subsection{Graph Neural Networks}
As we mentioned above, according to the assumptions on graph structure, Graph Neural Networks (GNNs) can be divided into Homogeneous Graph Neural Networks (HomoGNNs) and Heterogeneous Graph Neural Networks (HeteGNNs). Most HomoGNNs \cite{DBLP:conf/iclr/KipfW17,hamilton2017inductive,velivckovic2017graph} and HeteGNNs \cite{fu2020magnn,zhao2021heterogeneous,xu2021topic} are based on the message passing strategy \cite{gilmer2017neural}, in which each node gathers and transforms the messages from neighbours and then aggregates these messages to update the hidden state of the node. It is obvious that the working principle of GNNs is highly correlated with the structure of the input graph, which is fundamental to performing model inversion attacks on GNNs.

\subsection{Model Inversion Attacks on Graph Structured Data}
Thus far, only a very few model inversion attack methods have focused on the graph domain. Among these methods, GraphMI \cite{graphmi} was the first to be proposed, which can reconstruct the private graph in a white-box scenario by minimizing the cross-entropy loss obtained by feeding the intermediate reconstructed graph into the victim HomoGNN. A Feature-smoothness module was also deployed in GraphMI to make the connected nodes have similar features, which actually is an expression of the $1^{st}$ order proximity. GraphMI also contains an encoder-decoder module whose weights are drawn directly from the target white-box HomoGNN. Note that GraphMI is a white-box method. 

There is also a black-box version of GraphMI, called RL-GraphMI \cite{graphmiTrans}, which uses gradient estimation and reinforcement learning to achieve model inversion attack in black-box \cite{10098012} scenarios. Besides GraphMI and RL-GraphMI, the only graph model inversion attack method is GSEF \cite{DBLP:journals/popets/OlatunjiRFK23}, which is a post-hoc explainability techniques-based white-box attack. Both of these methods we mentioned above are focused on HomoGNNs, yet this paper is the first attempt to perform model inversion attack on HeteGNNs.

\section{Model Inversion Attack on HomoGNNs}
\label{sec:homoGNN}
\subsection{Preliminary}\label{Problem Statement}
In this paper, the goal of our attacker is to reconstruct the adjacency matrix of the original graph, different from \cite{graphmi}, our attack was conducted in a \textbf{gray-box} setup. That is, our attacker can not access the inner structures, hyper-parameters and the weights of the target GNN, but can access the logits of the fed graph to compute the cross-entropy loss. We also assume that the attacker has some auxiliary knowledge like \cite{graphmi} to facilitate model inversions, such as node labels $Y$, node features $X$, and edge density in the original training graph. Note that in this paper, we assume that the target GNN is designed for the node classification task.

Our goal is to reconstruct the private training adjacency matrix $A'$, which can maximize the similarity between $A'$ and the private training graph $A$. However, for our attacker, $A$ is unknowable. Thus we need to rely on some prior knowledge to infer the properties of the $A$ in a non-directive way. As shown in Fig. \ref{MIA}, we advocate that the reconstructed $A'$ should fall in the intersection of real-world data space \textbf{I} and model-covered space \textbf{II}. This means that $A'$ should have some shared proprieties to real-world graphs, i.e. $1^{st}$ and $2^{nd}$-order proximities and the target GNN should give us a higher accuracy when we feed $A'$ into this GNN since our intuition tell us that the $A'$ will be similar to the $A$ if the loss between true labels and predicted labels is minimized.

\subsection{Reconstruct Private Graph by Exploit Model Predicts}
As we motioned above, if the $A'$ is similar to $A$, we can expect a high classification performance of the target model when the $A'$ is fed. Following this intuition, we can optimize the reconstructed adjacency matrix by minimizing the training cross-entropy loss. Thus, our first objective function is:
\begin{equation}
\label{l_tar}
    \mathcal{L}_{tar} = - \sum_{i=1}^{|\mathcal{V}|} \sum_{k=1}^{|F|} Y_{ik} ln(Y_{ik}'),
\end{equation}
where $Y_{i k}$ denote the label matrix, $Y_{i k}'$ denote the prediction labels output from the target GNN $f(A', X)$. $Y_{i k}=1$ ($Y_{i k}'=1$) if node $v_i$ belong to class $k$.

In practice, given an undirected graph and its symmetric reconstructed adjacency matrix, in order to reduce the difficulty of the optimization problem, we only optimize the flattened upper triangular part of the $A'$, denoted as $b' \in \mathbb{R}^{|\mathcal{V}|(|\mathcal{V}|-1) / 2}$. We also relax $b' \in \{0,1\}^n$ to $b' \in [0,1]^n$ to reduce the difficulty caused by the discrete property of the adjacency matrix, in order to have an intuitive illustration, in what follows, we still use $A'$ instead of $b'$. Note that $A'$ and $b'$ can be converted to each other easily by execute the projection gradient method in \cite{graphmi}.

\subsection{Reconstruct Private Graph via $1^{st}$ and $2^{nd}$-order Proximities}\label{Graph-based similarity on HomoGs}
Assuming that the original graph $A$ is a real-world graph, we can make the reconstructed graph as similar to the original graph by improving the $1^{st}$ order proximity score and the $2^{nd}$ order proximity score. Note that this assumption can be easily satisfied since it is less meaningful to train a GNN on a fantasy graph.

The $1^{st}$ order proximity models the local similarity in graphs, which means that connected nodes are likely to have similar properties. Given a node pair $v_i$ and $v_j$, we model the $1^{st}$ order proximity on HomoGs between $v_i$ and $v_j$ as follows:
\begin{equation}
    p_{1^{st}}^{homo}(v_i, v_j) = A'_{ij}||x_i - x_j||_2^2,
\end{equation}
where $x_i \in \mathbb{R}^d$ is the $d$-dimension feature vector of node $v_i$ in $A'$, which can be extracted from $X$, and $||\cdot||_2$ is the $\ell_2$ norm. To force the reconstructed graph obey the $1^{st}$ order proximity, we minimize the following objective function:
\begin{equation}
    \mathcal{L}_{1^{st}}^{homo} = \sum_{i,j=1}^{|\mathcal{V}|}A'_{ij}||x_i - x_j||_2^2.
\end{equation}
We can get a simpler form for $\mathcal{L}_{1^{st}}$ via re-form it in a spectral perspective, that is:
\begin{equation}
    \mathcal{L}_{1^{st}}^{homo} = tr(X^{T}L'X),
\end{equation}
where $L'$ is the Laplacian matrix of $A'$, obtained as ${L}'= D' - A'$, and $X$ is the feature matrix. $D'$ is the diagonal degree matrix that can be obtained as $D'_{ii} = \sum_j A'_{ij}$.

The $2^{nd}$ order proximity is determined through the shared neighbourhood structure of nodes. Given the node $v_i$ and their one-hop neighbors $\mathcal{N}(v_i)$, we can model the $2^{nd}$ order proximity on HomoGs as follows:
\begin{equation}
    p_{2^{nd}}^{homo}(v_i, \mathcal{N}(v_i)) = ||x_i - \sum_{v_j \in \mathcal{N}(v_i)}x_j||_2^2.
\end{equation}
To force the reconstructed graph obey the $2^{nd}$ order proximities, we minimize the following objective function:
\begin{equation}
    \mathcal{L}_{2^{nd}}^{homo} = tr(X^TH'X),
\end{equation}
where $H' = (I-A')^T(I-A')$ is a symmetric matrix, $I$ is identity matrix.

Hence, the overall objective function of the proximities on HomoGs can be formulated as follows:
\begin{equation}
    \mathcal{L}_{pro}^{homo} = tr(X^T(L'+\beta H')X),
\end{equation}
where $\beta$ is a hyper-parameter. By minimizing $\mathcal{L}_{pro}^{homo}$, we can make the reconstructed graph have shared characteristics with the real-world graphs.

\subsection{Model Inversion Attack on HomoGNNs}
The overall objective function of model inversion attack on HomoGNNs is:
\begin{equation}\label{homooverall}
    \mathcal{L}_{homo} = \mathcal{L}_{tar} + \alpha \mathcal{L}_{pro}^{homo} + \gamma ||A'||_2,
\end{equation}
where $\alpha$, $\gamma$ is the hyper-parameter. $||A'||_2$ controls the sparsity of the reconstructed graph.

\begin{algorithm}[!ht]
	\caption{HomoGMI}
	\label{algorithm:HomoGMI}
	\LinesNumbered
	\KwIn {target model $f$, ground-truth label matrix $Y$, feature matrix $X$, iterations $\tau$, learning rate $\epsilon$, hyper-parameters $\alpha$, $\beta$ and $\gamma$}
	\KwOut {reconstructed graph $\mathcal{G'}$}
    Let $\lambda = 1$ \\
    Randomly initialize $A^{(\lambda)}$ \\
	\While{$\lambda \leq \tau$}{
        feed $A^{(\lambda)}, X$ into $f$ to get logits $X$ and predicted label $Y'$ \\
        compute $\mathcal{L}_{tar}$ via Eq. (\ref{l_tar}) \\        
    	compute $\mathcal{L}_{homo}$ via Eq. (\ref{homooverall}) \\
        update $A^{(\lambda)}$ to $A^{(\lambda+1)}$ via Eq. (\ref{homoupdate}) \\
        $t$ ++ \\
	}
    build $\mathcal{G}'=\{A^\tau, X\}$ \\
    \Return{$\mathcal{G}'$}
\end{algorithm}

Recall that we have relaxed $A' \in \{0,1\}^n$ to $A' \in [0,1]^n$. In order to turn the already relaxed $A'$ back into a discrete form, inspired by \cite{graphmi}, a projected gradient descent method is adopted. Mathematically, the $\lambda_{th}$ iteration of our optimization process can be described as follows:
\begin{equation}\label{homoupdate}
    A^{(\lambda+1)} = P(A^{(\lambda)} - \epsilon \ \nabla_{A^{(\lambda)}} \mathcal{L}_{homo}),
\end{equation}
where the initial $A^{(1)}$ is randomly initialized. $P(z)$ is a element-wise projection operation, $P(z)=0$ if $z<0$, $P(z)=1$ if $z>1$ and $P(z)=z$ otherwise. 

We call this model inversion attack method against HomoGNNs as HomoGMI. HomoGMI can also be described by the Algorithm \ref{algorithm:HomoGMI}.

\section{Model Inversion Attack on HeteGNNs}
\label{sec:heteGNN}
\subsection{Preliminary}
In this section, we extend our proposed HomoGMI on HeteGNNs, and we call this extended version as HeteGMI, which has a shared core idea with HomoGMI. However, in HeteGs, edges often associate different types of nodes, which is very different from HomoGs, thus, our  $1^{st}$ and $2^{nd}$-order proximities that are defined on HomoGs need to be modified. 

Recall that given a symmetric meta-path $m$, a symmetric meta-path augmented adjacency matrix $W^m$ can be carried out, and a homogeneous sub-graph can be generated from the $W^m$. As shown in Fig. \ref{graph_example_mi}, we can generate a homogeneous sub-graph from $W^{APA}$, and in this sub-graph, we can define the meta-path-based proximity on HeteGs according to this symmetric $W^{m}$. 

\subsection{Reconstruct Private Graph via Meta-path based Proximity}
The meta-path based $1^{st}$ order proximity models the local proximity in HeteGs, which means that the nodes connected via meta-path instances are similar. Mathematically, in the model inversion attack on HeteGNNs, our $1^{st}$ order proximity-based objective function can be expressed as:
\begin{equation}
    \mathcal{L}_{1^{st}}^{m} = \sum_{i, j = 1}^{|\mathcal{V}|} w_{ij}^{m'} || x_i - x_j ||_{2}^{2}.
\end{equation}
where $w_{ij}^{m'}$ indicates how many $m$-typed meta-path instances exist within $v_i$ and $v_j$ in $W^{m'}$. Note that in HeteGMI, the chosen of $m$ is not to be same to the target model, our experiments illustrate that an intuitive meta-path design is sufficient to make HeteGMI get a promising startup.

The meta-path based $2^{nd}$ order proximity is determined through the shared neighbourhood structure of nodes. It can be expressed as:
\begin{equation}
    \mathcal{L}_{2^{nd}}^{m} = \sum_{v_i \in \mathcal{V}}|| x_i - \sum_{v_j \in \mathcal{N}(v_i)^m} w_{ij}^{m'} x_j ||_{2}^{2},
\end{equation}
where $\mathcal{N}(v_i)^m$ denote the neighbors of node $v_i$ under the meta-path $m$. With the spectral perspective as same as we mentioned in Sec. \ref{Graph-based similarity on HomoGs}, we re-form the $\mathcal{L}_{1^{st}}^{m}$ and $\mathcal{L}_{2^{nd}}^{m}$ as follows:
\begin{equation}
\begin{aligned}
    \mathcal{L}_{1^{st}}^{m} &= 2tr(X^T L^{m'} X), \\
    \mathcal{L}_{2^{nd}}^{m} &= 2tr(X^T H^{m'} X),
\end{aligned}
\end{equation}
where $H^{m'} = (I - W^{m'})^T (I - W^{m'})$  is symmetric.

For all $t$ symmetric meta-paths ${m^1, m^2, \dots, m^t}$, we fuse them together, that is $W = \sum_{i=1}^{t} W^{m^t}$, $D = \sum_{i=1}^{t} D^{m^t}$. Thus, the $\mathcal{L}_{pro}$ we defined in HeteGs can be described as:
\begin{equation}
    \mathcal{L}_{pro}^{hete} = tr(X^T (L' + \beta H') X),
\end{equation}
where $H' = (I - W')^T (I - W')$, $L' = D' - W'$.

\subsection{Model Inversion Attack on HeteGNNs}
The overall objective function of a model inversion attack on HeteGNNs is:
\begin{equation}\label{heteoverall}
    \mathcal{L}_{hete} = \mathcal{L}_{tar} + \alpha \mathcal{L}_{pro}^{hete}  + \gamma ||A'||_2,
\end{equation}
where $\alpha$, $\gamma$ is the hyper-parameter. $||a||_2$ controls the sparsity of the reconstructed graph. $A'$ is the full adjacency matrix combined from $\{A^{c_1}, $ $A^{c_2}, \dots, A^{c_K}\}$. 

The $\lambda_{th}$ iteration our
optimization process for each $A^{c_k} \in \{A^{c_1}, $ $A^{c_2}, \dots, A^{c_K}\}$, which the processes can be described as follows:
\begin{equation}\label{heteupdate}
    ({A^{c_k}})^{(\lambda+1)} = P(({A^{c_k}})^{(\lambda)} + \epsilon \ \nabla_{({A^{c_k}})^{(\lambda)}} \mathcal{L}_{hete}),
\end{equation}
where $P$ is the projection operator that we used in Eq. \ref{homoupdate}.

We call this model inversion attack method on HeteGs as HeteGMI. In HeteGMI, given a HeteG with $K$ edge types, our strategy is to optimize the relaxed edge-type specified matrices $\{{A^{c_1}}', {A^{c_2}}', \dots, {A^{c_K}}'\}$ directly, and then combine them into a full-sized adjacency matrix $A'$ to compute $\mathcal{L}_{tar}$. Then, given $t$ symmetric meta-paths $\mathcal{M} = \{m^1, m^2, \dots, m^t\}$, we compute their corresponding meta-path augmented adjacency matrices $\{{W^{m^1}}', {W^{m^2}}', \dots, {W^{m^t}}'\}$, and evaluate the meta-path based similarity for each meta-path augmented adjacency matrix, which the similarity will be introduced in the following subsection. Note that in our proposed methods, these symmetric meta-paths $\mathcal{M}$ need to be defined by the attacker manually.

HeteGMI can also be described by the Algorithm \ref{algorithm:HeteGMI}.
\begin{algorithm}[!ht]
	\caption{HeteGMI}
	\label{algorithm:HeteGMI}
	\LinesNumbered
	\KwIn {target model $f$, label vector $Y$, feature matrix $X$, iterations $\tau$, learning rate $\epsilon$, edge types $\mathcal{C} = \{c_1, c_2, \dots, c_K\}$, symmetric meta-paths $\mathcal{M}=\{m^1, m^2, \dots, m^t\}$, hyper-parameters $\alpha$, $\beta$ and $\gamma$}
	\KwOut {reconstructed graph $\mathcal{G'}$}    
    Let $\lambda = 1$ \\
    \For {$c_k \in \mathcal{C}$} {
            Randomly initialize $(A^{c_k})^\lambda$
    }
	\While{$\lambda \leq \tau$}{
        combine $\{(A^{c_1})^{(\lambda)}, (A^{c_2})^{(\lambda)}, \dots, (A^{c_K})^{(\lambda)}\}$ to the full adjacency matrix $A^{(\lambda)}$ \\
        feed $A^{(\lambda)}, X$ into $f$ to get logits $X$ and predicted label $Y'$ \\    
        compute $\mathcal{L}_{tar}$ via Eq. (\ref{l_tar}) \\
        \For {$m \in \mathcal{M}$} {
            compute $W^m$ according to Eq. (\ref{computew}) \\
        }
    	compute $\mathcal{L}_{hete}$ via Eq. (\ref{heteoverall}) \\  
        \For {$c_k \in \mathcal{C}$} {
            update $(A^{c_k})^{(\lambda)}$ to $(A^{c_k})^{(\lambda+1)}$ via Eq. (\ref{heteupdate}) \\
        }
        $\lambda$ ++ \\
	}
    combine $\{(A^{c_1})^\tau, (A^{c_2})^\tau, \dots, (A^{c_K})^\tau\}$ to the full adjacency matrix $A^\tau$ \\
    build $\mathcal{G}' = \{A^\tau, X\}$\\
    \Return{$\mathcal{G}'$}
\end{algorithm}

\section{Experiments}
\label{sec:experiments}
In this section, we present the experimental results to show the effectiveness of the proposed HomoGMI and HeteGMI.

\subsection{Experimental Settings}
\subsubsection{Datasets}
For HomoGMI, we adopt three real-world HomoG datasets: Cora, Cora-ML and Citeseer. And we adopt other three real-world HeteG datasets ACM, DBLP and IMDB for HeteGMI. Specifically, ACM contains three types of nodes (including papers(P), authors(A), and subjects (S)) and two edge types (including PA and PS). DBLP contains three types of nodes (including papers(P), authors(A) and conferences (C)) and two edge types (including PA and PC). IMDB contains three types of nodes (including movies(M), actors(A), and directors(D)) and two edge types (including MA and DM). The overview of these datasets is shown in Table \ref{dataset_table}.

\begin{table}[ht]
	\renewcommand\arraystretch{1}
	\caption{The overview of the datasets.}
	\label{dataset_table}
	\centering
	\begin{tabular}{cccccc}
		\toprule
		Dataset & \#Nodes & \#Edges & \#Node type & \#Edge type & Meta-paths \\ \midrule
            Cora    & 2708    & 5278   & -           & -          & -            \\
            Cora-ML    & 2995    & 8158   & -           & -        & -              \\
            Citeseer    & 3312    & 4536   & -           & -       & -               \\ \midrule      
            \midrule            
		ACM     & 8994    & 12961   & 3           & 2        & PAP, PSP              \\
  		DBLP    & 18,405    & 67,946   & 3           & 2         & APCPA, APA             \\
		IMDB    & 12,772   & 18,644   & 3           & 2       & MAM, MDM               \\
		\bottomrule
	\end{tabular}
\end{table}

\begin{table}[h]
    \centering
    \caption{The accuracy (\%) of target HomoGNNs.}
    \label{accHomoGNNs}
    \small
    \setlength\tabcolsep{4pt}
    \renewcommand{\arraystretch}{1}
    \begin{tabular}{c|cl|cl|cl}
        \toprule
        Datasets  & \multicolumn{2}{c}{Cora}  & \multicolumn{2}{c}{Cora-ML} & \multicolumn{2}{c}{Citeseer} \\ 
        \midrule
        GCN       & \multicolumn{2}{c}{79.30} & \multicolumn{2}{c}{83.70}   & \multicolumn{2}{c}{73.49}      \\
        GraphSAGE & \multicolumn{2}{c}{76.90} & \multicolumn{2}{c}{79.90}   & \multicolumn{2}{c}{63.10}     \\
        \bottomrule
    \end{tabular}
\end{table}

\begin{table}[h]
    \centering
    \caption{The micro and macro F1-score (\%) of target HeteGNNs.}
    \label{accHeteGNNs}
    \renewcommand{\arraystretch}{1}
    \begin{tabular}{c|cc|cc|cc}
    \toprule
    Datasets & \multicolumn{2}{c|}{ACM} & \multicolumn{2}{c|}{DBLP} & \multicolumn{2}{c}{IMDB} \\
    \midrule
    Metrics  & Micro-F1   & Macro-F1   & Micro-F1    & Macro-F1   & Micro-F1    & Macro-F1   \\
    \midrule
    HAN      & 90.79      & 90.89      & 92.17       & 91.67      & 59.80       & 57.74      \\
    GTN      & 91.96      & 91.31      & 93.97       & 93.52      & 61.02       & 60.47      \\
    RGCN     & 91.41      & 91.55      & 93.16       & 91.52      & 62.05       & 58.85     \\
    \bottomrule
    \end{tabular}
\end{table}

\begin{table}[ht]
\renewcommand\arraystretch{1.0}
\centering
\small
\caption{Results of model inversion attack on GCN and GraphSAGE (\%). Higher is better.}
\label{tab:mainreshomo}
\begin{tabular}{c|c|cc|cc|cc}
    \toprule
    \multirow{2}{*}{Target Model} & \multirow{2}{*}{Attack Method} & \multicolumn{2}{c|}{Cora}       & \multicolumn{2}{c|}{Cora-ML}    & \multicolumn{2}{c}{Citeseer} \\
                                  &                                & AUC            & AP             & AUC            & AP             & AUC            & AP                     \\ \midrule
    \multirow{4}{*}{GCN}          & Sim-Attr                           & 78.91          & 81.07          & 75.58          & 82.90          & 87.00          & 85.42                 \\
                                  & Sim-Emb                            & 55.90          & 55.18          & 56.39          & 55.52          & 58.61          & 58.77              \\
                                  & GraphMI                        & 86.80          & 88.30          & 86.68          & 85.13          & 87.80          & \textbf{88.50}          \\
                                  & \textbf{HomoGMI}                           & \textbf{89.18} & \textbf{89.35} & \textbf{88.81} & \textbf{86.88} & \textbf{90.26} & 87.58           \\ \midrule
    \multirow{4}{*}{GraphSAGE}   & Sim-Attr                           & 78.03          & 80.43          & 82.46          & 85.93          & 86.08          & 84.51                 \\
                                  & Sim-Emb                            & 55.88          & 55.32          & 56.02          & 56.25          & 57.64          & 57.81                  \\
                                  & GraphMI                        & 79.26          & 74.72          & 88.88          & \textbf{89.64} & 89.74          & 90.98                 \\
                                  & \textbf{HomoGMI}                           & \textbf{87.01} & \textbf{84.38} & \textbf{89.57} & 86.22          & \textbf{90.63} & \textbf{91.55}  \\ \bottomrule
\end{tabular}
\end{table}

\subsubsection{Target Model}
For HomoGMI, we adopt two HomoGNNs: GCN \cite{DBLP:conf/iclr/KipfW17} and GraphSAGE \cite{hamilton2017inductive} and for HeteGMI, we adopt three HeteGNNs: HAN \cite{wang2019heterogeneous}, GTN \cite{hu2020heterogeneous} and RGCN \cite{schlichtkrull2018modeling}. The performance of our target model has shown in Table \ref{accHomoGNNs} and Table \ref{accHeteGNNs}.

\subsubsection{Baselines.}
We adopt three Baselines: Sim-Attr, Sim-Emb and GraphMI.
\begin{itemize}[leftmargin=*]
    \item Sim-Attr is measured by cosine distance among node attributes, and the higher similarity typically indicates a higher probability of an edge exists within any two nodes.
    \item Sim-Emb is similar to the Sim-Attr. The only difference is Sim-Emb is measured by cosine distance among the extracted node embeddings from the penultimate layer of the target GNNs.
    \item We note that GraphMI offers many similarities with our methods. The main difference is that our methods use $1^{st}$ and $2^{nd}$ order proximity to make the generated graph similar to real-world graphs. In addition, the graph is then reconstructed by GraphGMI using the graph auto-encoder module, in which the encoder is replaced by learned parameters of the target model, and the decoder is a logistic function.
\end{itemize}

Considering that there does not exist a model inversion method that is applicable to HeteGNNs so far, thus, a different approach was taken to evaluate the performance of HeteGMI: \textbf{(i)} For our proposed HeteGMI and Sim-Emb, we require them to reconstruct a complete HeteG first and then extract the homogeneous sub-graph derived from the symmetrical meta-paths we defined in Table \ref{dataset_table}. \textbf{(ii)} For Sim-Attr and GraphMI, since some node types on the HeteG do not have features and labels, hence, we only require them to reconstruct the homogeneous sub-graph directly as the same as the extracted sub-graph by the meta-paths we defined in Table \ref{dataset_table}.

\subsubsection{Metrics}
Since our attack is unsupervised, the attacker cannot find a threshold to make a concrete prediction through the algorithm. To evaluate our attack, we use AUC (area under the ROC curve) and AP (average precision) as our metrics, which is consistent with previous model inversion attack works \cite{graphmi}. In experiments, we use all the edges from the training graph and the same number of randomly sampled pairs of unconnected nodes (non-edges) to evaluate AUC and AP.

\subsection{Quantitative Evaluation}
We first evaluate the performance of HomoGMI, and the results are shown in Table \ref{tab:mainreshomo}, from which we can get the following observations.
\begin{itemize}
    \item In most cases, HomoGMI achieves better results than GraphMI \cite{graphmi}. This is because GraphMI ignores the $1^{st}$ and $2^{nd}$ order proximity on the graph-structured data.
    \item Another observation from Table \ref{tab:mainreshomo} is that there is not a clear connection between the accuracy of the model and the final inversion performance. In our experimental results shown in Table \ref{accHomoGNNs}, GCN reported 87.03\% classification accuracy on the Cora dataset and GraphSAGE reported 76.90\% classification accuracy on this dataset, yet the attack performance did not show a significant margin between GCN and GraphSAGE.
\end{itemize}

The permanence of the proposed HeteGMI was reported in Table \ref{tab:mainreshete} and from which we can get the following observations: 
\begin{itemize}[leftmargin=*]
    \item In most cases, HomoGMI achieves better results than the baseline methods, and it seems that our proposed HeteGMI seems to be the only method that works, since the AUC and AP of the other methods are around 50, which suggests that they do not work well on the HeteGNNs, although we only ask them to reconstruct the homogeneous sub-graph of the HeteG. However, we can also observe that compared to the AUC and AP values of around 90 on HomoGs, there is still considerable progress to be made for model inversion attacks on HeteGs.
    \item HeteGMI performs worse on those HeteGNNs that can automatically extract meta-paths (i.e., GTN) than those that can not extract the meta-path automatically (i.e., HAN and RGCN). This may be because the meta-paths extracted within the GTN are not stable during our model inversion attack process, making the final output of GTN is also unstable, which causes difficulties for the optimization of $\mathcal{L}_{tar}$.
    \item The performance of HeteGMI is highly related to the target model's performance. For example, the attack performance on IMDB is notably lower than other datasets, while in Table \ref{accHeteGNNs}, we can see that the classification performance on the IMDB dataset is also notably lower than other datasets. 
\end{itemize}

\begin{table}[ht]
\renewcommand\arraystretch{1.0}
\centering
\caption{Results of model inversion attack on HAN, GTN and RGCN (\%). Higher is better.}
\label{tab:mainreshete}
\begin{tabular}{c|c|cc|cc|cc}
\toprule
\multirow{2}{*}{Target Model} & \multirow{2}{*}{Attack Method} & \multicolumn{2}{c|}{ACM}                   & \multicolumn{2}{c|}{DBLP}       & \multicolumn{2}{c}{IMDB}       \\
                              &                                & AUC                       & AP             & AUC            & AP             & AUC            & AP             \\ \midrule
\multirow{4}{*}{HAN}          & Sim-Attr                       & \multicolumn{1}{r}{53.20} & 53.20          & 52.35          & 52.34          & 56.61          & 56.61          \\
                              & Sim-Emb                        & 55.32                     & 55.32          & 54.38          & 54.38          & 62.46          & 62.46          \\
                              & GraphMI                        & 52.75                     & 56.02          & 46.41          & 64.81          & 37.95          & 48.63          \\
                              & \textbf{HeteGMI}               & \textbf{75.75}            & \textbf{73.45} & \textbf{84.79} & \textbf{84.31} & \textbf{67.34} & \textbf{65.29} \\ \midrule
\multirow{4}{*}{GTN}          & Sim-Attr                       & 53.20                     & 53.20          & 52.35          & 52.34          & 56.61          & 56.61          \\
                              & Sim-Emb                        & 54.47                     & 54.46          & 66.25           & 66.25         & 62.27          & 62.27          \\
                              & GraphMI                        & 50.90                     & 54.48          & 39.46           & 57.36         & 37.47          & 48.21          \\
                              & \textbf{HeteGMI}               & \textbf{73.56}            & \textbf{69.77} & \textbf{75.20}   & \textbf{75.89}     & \textbf{65.29} & \textbf{63.63} \\ \midrule
\multirow{4}{*}{RGCN}         & Sim-Attr                       & 53.20                     & 53.20          & 52.35          & 52.34          & 56.61          & 56.61          \\
                              & Sim-Emb                        & 54.55                     & 54.55          & 54.49          & 54.49          & 61.83          & 61.83          \\
                              & GraphMI                        & 50.95                     & 54.06          & 41.41          & 59.81          & 36.31          & 47.63          \\
                              & \textbf{HeteGMI}               & \textbf{73.60}            & \textbf{70.42} & \textbf{86.78} & \textbf{85.52} & \textbf{67.37} & \textbf{67.38} \\ \bottomrule
\end{tabular}
\end{table}


\subsection{Ablation Studies}
In order to better validate the effectiveness of the proposed method, we perform ablation for each component of the proposed method. Basically, our method is an optimization method, so we focus on the proposed optimization objectives, which for both HomoGMI and HeteGMI contain four parts: $\mathcal{L}_{tar}$, $\mathcal{L}_{pro}$ (including $\mathcal{L}_{1^{st}}$ and $\mathcal{L}_{2^{nd}}$), and the $||A'||$ term. Thus, for a complete evaluation, we ablate all four optimization objectives for both HomoGMI and HeteGMI. The results are listed in Table \ref{tab:AblationHOMO}, which shows that after removing these four parts from HomoGMI and HeteGMI, the attack performance has also been reduced in different magnitudes and shows the effect of our four optimization objectives. We can also observe that $\mathcal{L}_{tar}$ may be the most important part that guarantees an acceptable model inversion performance. This phenomenon is similar to the model inversion attack on image data.

\begin{table}[ht]
\renewcommand\arraystretch{1.0}
\centering
\small
\caption{Ablation Studies of HomoGMI and HeteGMI.}
\label{tab:AblationHOMO}
\begin{tabular}{c|c|cc|cc|cc|cc|cc}
\toprule
                      \multirow{2}{*}{Target Model} & \multirow{2}{*}{Datasets}  & \multicolumn{2}{c|}{FULL} & \multicolumn{2}{c|}{- $\mathcal{L}_{tar}$} & \multicolumn{2}{c|}{- $\mathcal{L}_{1^{st}}$} & \multicolumn{2}{c|}{- $\mathcal{L}_{2^{nd}}$} & \multicolumn{2}{c}{- $||A'||$} \\
                      &           & AUC         & AP         & AUC           & AP           & AUC             & AP              & AUC             & AP              & AUC               & AP                \\
                      \midrule
\multirow{4}{*}{GCN}  & Cora      & \textbf{89.18}       & \textbf{89.35}      & 78.48         & 86.52        & 83.53           & 85.12           & 83.70           & 81.87           & 85.44             & 85.65             \\
                      & Cora-ML   & \textbf{88.81}       & \textbf{86.88}      & 84.01         & 83.86        & 87.13           & 84.95           & 86.88           & 85.19           & 87.00             & 87.83             \\
                      & Citeseer  & \textbf{90.26}       & \textbf{87.58}      & 81.11         & 81.16        & 88.35           & 86.24           & 86.94           & 84.14           & 89.59             & 88.62             \\
                      \midrule
\multirow{4}{*}{GraphSAGE} & Cora      & \textbf{87.01}       & \textbf{84.38}      & 76.70         & 72.10        & 86.37           & 78.35           & 81.25           & 78.94           & 85.49             & 85.71             \\
                      & Cora-ML   & \textbf{89.57}       & \textbf{86.22}      & 82.02         & 82.58        & 84.76           & 81.99           & 87.97           & 84.56           & 86.26             & 84.17             \\
                      & Citeseer  & \textbf{90.63}       & \textbf{91.55}      & 76.65         & 77.24        & 84.88           & 81.22           & 86.12           & 86.15           & 82.79             & 86.55             \\
                      \midrule    
                      \midrule
\multirow{3}{*}{HAN}          & ACM                       & \textbf{75.75}       & \textbf{73.45}      & 50.16         & 50.16        & 72.20           & 66.18           & 73.18           & 67.02           & 72.00             & 67.51             \\
                              & DBLP                      & \textbf{84.79}       & \textbf{84.31}      & 51.11         & 51.11        & 83.89           & 83.09           & 83.90           & 83.10           & 83.84             & 83.31             \\
                              & IMDB                      & \textbf{67.34}       & \textbf{65.29}      & 50.19         & 50.19        & 65.19           & 64.03           & 65.06           & 62.96           & 64.32             & 62.98             \\
                              \midrule
\multirow{3}{*}{GTN}          & ACM                       & \textbf{73.56}       & \textbf{69.77}      & 50.01         & 50.01        & 72.62           & 67.89           & 71.35           & 67.61           & 70.56             & 66.48             \\
                              & DBLP                      & \textbf{75.20}       & \textbf{75.89}      & 51.11         & 51.11        & 70.65           & 73.41           & 70.99           & 73.41           & 70.64             & 73.41             \\
                              & IMDB                      & \textbf{65.29}       & \textbf{63.63}      & 50.19         & 50.19        & 65.22           & 63.37           & 65.72           & 63.13           & 63.72             & 62.86             \\
                              \midrule
\multirow{3}{*}{RGCN}         & ACM                       & \textbf{73.60}       & \textbf{70.42}      & 50.24         & 50.24        & 72.62           & 68.81           & 72.62           & 67.89           & 72.62             & 67.89             \\
                              & DBLP                      & \textbf{86.78}       & \textbf{85.52}      & 50.94         & 50.94        & 85.96           & 84.92           & 85.21           & 85.19           & 85.67             & 84.32             \\
                              & IMDB                      & \textbf{67.37}       & \textbf{67.38}      & 50.19         & 50.19        & 65.77           & 64.80           & 65.56           & 62.81           & 63.13             & 60.65           \\ 
\bottomrule
\end{tabular}
\end{table}

\subsection{Hyper-parameter Studies}
\begin{figure} [!ht]
	\centering
    \includegraphics[width=\textwidth]{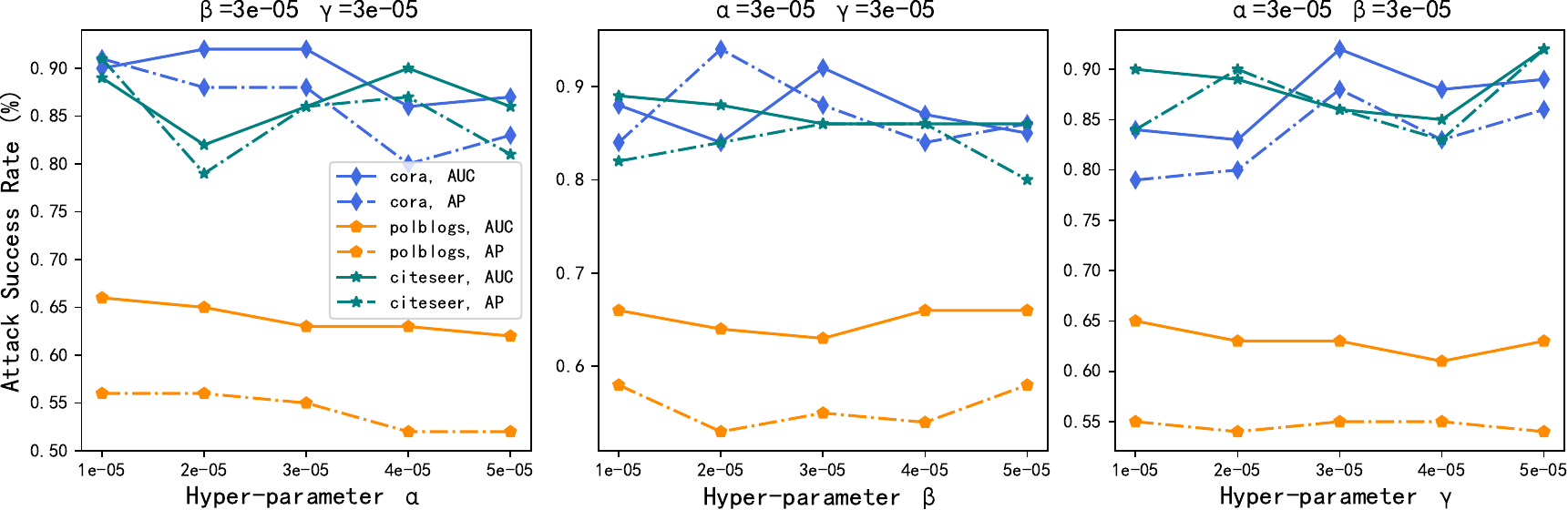}\\
    \includegraphics[width=\textwidth]{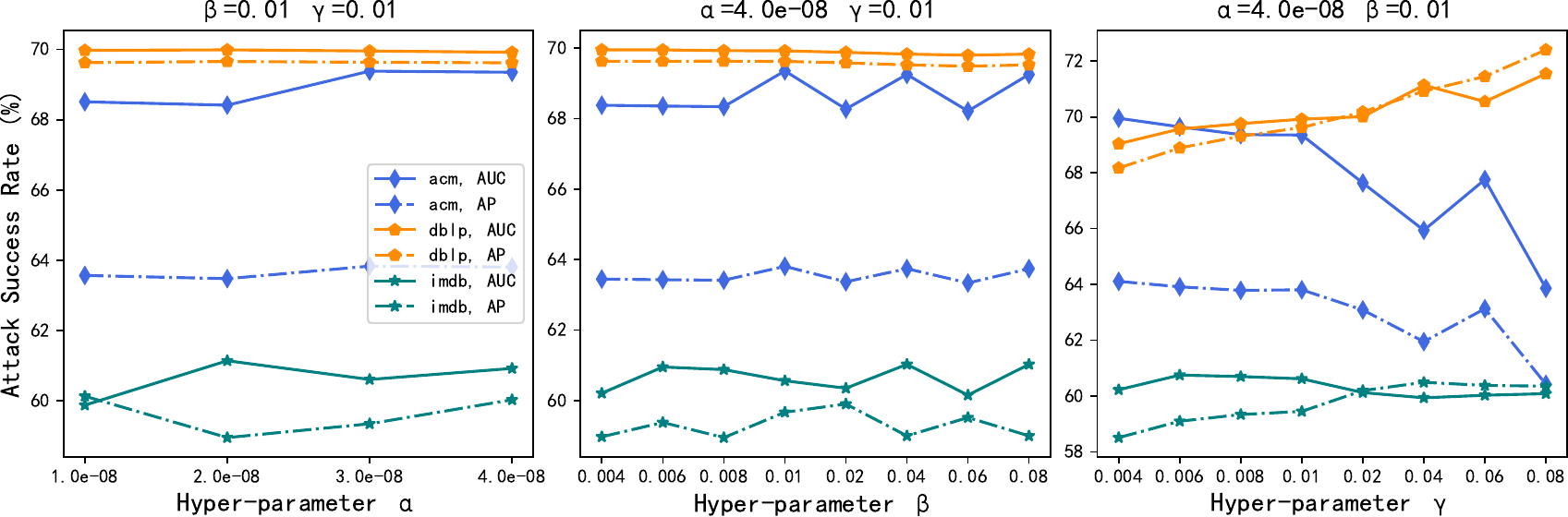} 
	\caption{Hyper-parameter study on HomoGMI (top) and HeteGMI (bottom).}
	\label{fig:parameter} 
\end{figure}

In this subsection, we evaluate the impact of the five hyper-parameters: $\alpha, \beta, \gamma$ (see Eq. \ref{homooverall} and Eq. \ref{heteoverall}). In which, $\alpha$ is used to control the weight of graph structure similarity loss term in the overall loss, $\beta$ is used to reconcile the proportion of first-order proximity and second-order proximity in the graph structure loss term, $\gamma$ is used to control the weight of $||A'||$ loss term in the overall loss. The results of the hyper-parameter studies on HomoGMI and HeteGMI are shown in Fig. \ref{fig:parameter}. We can see that our proposed HomoGMI and HeteGMI are basically stable with different parameter settings, the margin of the highest value and the lowest value is around 6\%. 

\subsection{Attack with different Meta-paths}
\begin{figure}[ht]
    \centering
    \includegraphics[width=\textwidth]{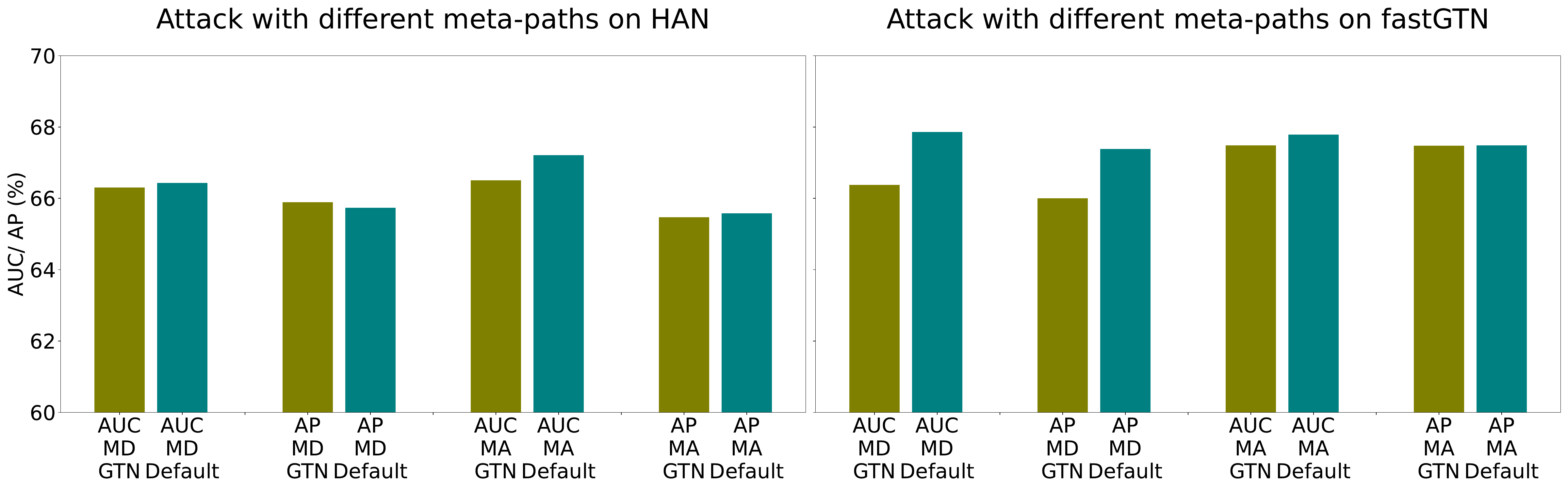} 
    \caption{The performance of HeteGMI with different mete-paths. The first row of the caption X-axis is a different metric (AUC or AP), the second row is different edge types (MD or MA), and the third row is the meta-path settings (GTN-learned or Default)}
    \label{fig:diff_meta}
\end{figure}

The proposed HeteGMI needs to define the meta-paths manually. Thus, it is also meaningful to evaluate the performance of HeteGMI on different meta-paths. Here we use the meta-paths on the IMDB dataset learned by GTN\cite{yun2019graph} as a comparison, i.e., MDM, MAM and MDMDM, as a counterpart, the default meta-paths are MAM and MDM. The results are shown in Fig. \ref{fig:diff_meta}, from which we can see that the effect of different meta-path settings on HeteGMI is slight, probably due to the wide availability of $1^{st}$ and $2^{nd}$ order proximities, which allows HeteGMI to be used in a variety of meta-path settings.

\begin{figure}[!ht]
    \centering
    \includegraphics[width=0.9\textwidth]{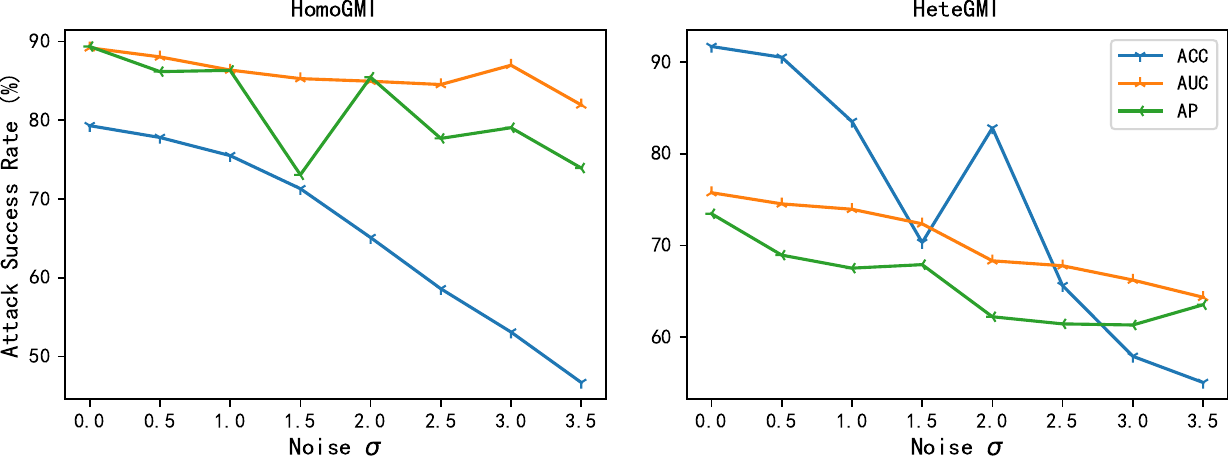}    
    \caption{The attack performance under different noises of HomoGMI and HeteGMI. The blue line denotes the classification accuracy of the victim model under different noises. The yellow and green lines denote the model inversion attack performance.}
    \label{fig:noise}
\end{figure}


\subsection{Attack with Random Noises}
In practice, the target victim model may equip with defence mechanisms, i.e., some random noises may be added to the returned results of this target model, which can be used to implement Differential Privacy mechanisms. In order to evaluate the attack capability of HomoGMI and HeteGMI in practice, we add $(\mu, \sigma^2)$ Gaussian noises to the target model return results in this subsection, where $\mu$ is fixed with $1$ and $\sigma \in [0.5, 3.5]$. In this experiment, the victim model for HomoGMI is GCN, and the dataset is Cora; the victim model for HeteGMI is HAN, and the dataset is ACM.

The results are shown in Fig. \ref{fig:noise}, and we can see that with the enhancement of noise, the target model becomes almost unusable. However, our proposed HomoGMI and HeteGMI still achieve acceptable attack performance, especially HomoGMI. This experimental result suggests that it is not wise to defend against model inversion attacks by adding noises to the returned results of the victim model.


\section{Conclusion}
\label{sec:conclusion}
Overarchingly, we propose an action guideline to perform the model inversion attack against Homogeneous Graph Neural Networks and Heterogeneous Graph Neural Networks from the perspective of the Model Covered Space and the Real World Data Space, providing a brand new insight into the privacy issue of Graph Neural Networks.

Notably, we consider both $1^{st}$ and $2^{nd}$ order proximities on graph-structured data and propose a new model inversion attack strategy named HomoGMI for Homogeneous Graph Neural Networks. Compared to existing methods, extensive experimental results on seven public datasets have shown the superiority of the proposed method with respect to several metrics. 

Furthermore, to fill the gap of model inversion attacks on Heterogeneous Graph Neural Networks, we propose HeteGMI, which utilises the Meta-path based proximity to make the reconstructed graph fill into the Real World Data Space. Our experiments reveal that the proposed HeteGMI outperforms all baselines. Notably, HeteGMI is the first attempt to implement model inversion attacks on Heterogeneous Graph Neural Networks.

\subsubsection{Acknowledgements} 
This work is supported in part by Yunnan Province Education Department Foundation under Grant No.2022j0008, in part by the National Natural Science Foundation of China under Grant 62162067 and 62101480, Research and Application of Object Detection based on Artificial Intelligence, in part by the Yunnan Province expert workstations under Grant 202205AF150145.

\bibliographystyle{splncs04}
\bibliography{ref}

\begin{thebibliography}{10}
\providecommand{\url}[1]{\texttt{#1}}
\providecommand{\urlprefix}{URL }
\providecommand{\doi}[1]{https://doi.org/#1}

\bibitem{s2v}
Dai, H., Dai, B., Song, L.: Discriminative embeddings of latent variable models
  for structured data. In: ICML (2016)

\bibitem{fan2019metapath}
Fan, S., Zhu, J., Han, X., Shi, C., Hu, L., Ma, B., Li, Y.: Metapath-guided
  heterogeneous graph neural network for intent recommendation. In: KDD (2019)

\bibitem{fu2020magnn}
Fu, X., Zhang, J., Meng, Z., King, I.: {MAGNN:} metapath aggregated graph
  neural network for heterogeneous graph embedding. In: {WWW} (2020)

\bibitem{gilmer2017neural}
Gilmer, J., Schoenholz, S.S., Riley, P.F., Vinyals, O., Dahl, G.E.: Neural
  message passing for quantum chemistry. In: {ICML} (2017)

\bibitem{hamilton2017inductive}
Hamilton, W.L., Ying, Z., Leskovec, J.: Inductive representation learning on
  large graphs. In: NeurIPS (2017)

\bibitem{hou2017hindroid}
Hou, S., Ye, Y., Song, Y., Abdulhayoglu, M.: Hindroid: An intelligent android
  malware detection system based on structured heterogeneous information
  network. In: KDD (2017)

\bibitem{hu2018leveraging}
Hu, B., Shi, C., Zhao, W.X., Yu, P.S.: Leveraging meta-path based context for
  top- {N} recommendation with {A} neural co-attention model. In: KDD (2018)

\bibitem{hu2020heterogeneous}
Hu, Z., Dong, Y., Wang, K., Sun, Y.: Heterogeneous graph transformer. In: WWW
  (2020)

\bibitem{DBLP:journals/comcom/Jiang22}
Jiang, W.: Graph-based deep learning for communication networks: {A} survey.
  Computer Communications  \textbf{185},  40--54 (2022)

\bibitem{DBLP:conf/iclr/KipfW17}
Kipf, T.N., Welling, M.: Semi-supervised classification with graph
  convolutional networks. In: ICLR (2017)

\bibitem{DBLP:journals/tmc/LeeYD23}
Lee, M., Yu, G., Dai, H.: Decentralized inference with graph neural networks in
  wireless communication systems. {IEEE} Transactions on Mobile Computing
  \textbf{22}(5),  2582--2598 (2023)

\bibitem{10098012}
Li, H., Zhang, J., Gao, S., Wu, L., Zhou, W., Wang, R.: Towards query-limited
  adversarial attacks on graph neural networks. In: ICTAI (2022)

\bibitem{liu2020heterogeneous}
Liu, S., Ounis, I., Macdonald, C., Meng, Z.: A heterogeneous graph neural model
  for cold-start recommendation. In: {SIGIR} (2020)

\bibitem{luo2020dynamic}
Luo, W., Zhang, H., Yang, X., Bo, L., Yang, X., Li, Z., Qie, X., Ye, J.:
  Dynamic heterogeneous graph neural network for real-time event prediction.
  In: KDD (2020)

\bibitem{DBLP:journals/corr/abs-2203-17209}
Montanari, A., Wu, Y.: Adversarial examples in random neural networks with
  general activations. CoRR  \textbf{abs/2203.17209} (2022)

\bibitem{DBLP:journals/popets/OlatunjiRFK23}
Olatunji, I.E., Rathee, M., Funke, T., Khosla, M.: Private graph extraction via
  feature explanations. Proceedings on Privacy Enhancing Technologies
  \textbf{2023}(2),  59--78 (2023)

\bibitem{ribeiro2017struc2vec}
Ribeiro, L.F., Saverese, P.H., Figueiredo, D.R.: struc2vec: Learning node
  representations from structural identity. In: KDD (2017)

\bibitem{schlichtkrull2018modeling}
Schlichtkrull, M.S., Kipf, T.N., Bloem, P., van~den Berg, R., Titov, I.,
  Welling, M.: Modeling relational data with graph convolutional networks. In:
  ESWC (2018)

\bibitem{tang2015line}
Tang, J., Qu, M., Wang, M., Zhang, M., Yan, J., Mei, Q.: Line: Large-scale
  information network embedding. In: WWW (2015)

\bibitem{line}
Tang, J., Qu, M., Wang, M., Zhang, M., Yan, J., Mei, Q.: Line: Large-scale
  information network embedding. In: WWW (2015)

\bibitem{velivckovic2017graph}
Veli{\v{c}}kovi{\'c}, P., Cucurull, G., Casanova, A., Romero, A., Lio, P.,
  Bengio, Y.: Graph attention networks. arXiv preprint  (2017)

\bibitem{ijcai2019-522}
Wang, S., Chen, Z., Yu, X., Li, D., Ni, J., Tang, L., Gui, J., Li, Z., Chen,
  H., Yu, P.S.: Heterogeneous graph matching networks for unknown malware
  detection. In: IJCAI (2019)

\bibitem{wang2019heterogeneous}
Wang, X., Ji, H., Shi, C., Wang, B., Ye, Y., Cui, P., Yu, P.S.: Heterogeneous
  graph attention network. In: WWW (2019)

\bibitem{MetaPathBasedProximity}
Wang, X., Lu, Y., Shi, C., Wang, R., Cui, P., Mou, S.: Dynamic heterogeneous
  information network embedding with meta-path based proximity. IEEE
  Transactions on Knowledge and Data Engineering  \textbf{34}(3),  1117--1132
  (2022)

\bibitem{xiang2022temporal}
Xiang, S., Cheng, D., Shang, C., Zhang, Y., Liang, Y.: Temporal and
  heterogeneous graph neural network for financial time series prediction. In:
  CIKM. pp. 3584--3593 (2022)

\bibitem{xu2021topic}
Xu, S., Yang, C., Shi, C., Fang, Y., Guo, Y., Yang, T., Zhang, L., Hu, M.:
  Topic-aware heterogeneous graph neural network for link prediction. In: CIKM
  (2021)

\bibitem{DBLP:journals/tnn/YuanHZL19}
Yuan, X., He, P., Zhu, Q., Li, X.: Adversarial examples: Attacks and defenses
  for deep learning. {IEEE} Trans. Neural Networks Learn. Syst.
  \textbf{30}(9),  2805--2824 (2019)

\bibitem{yun2019graph}
Yun, S., Jeong, M., Kim, R., Kang, J., Kim, H.J.: Graph transformer networks
  (2019)

\bibitem{graphmiTrans}
Zhang, Z., Liu, Q., Huang, Z., Wang, H., Lee, C.K., Chen, E.: Model inversion
  attacks against graph neural networks. IEEE Transactions on Knowledge and
  Data Engineering pp. 1--13 (2022)

\bibitem{graphmi}
Zhang, Z., Liu, Q., Huang, Z., Wang, H., Lu, C., Liu, C., Chen, E.: Graphmi:
  Extracting private graph data from graph neural networks. In: IJCAI (2021)

\bibitem{zhao2021heterogeneous}
Zhao, J., Wang, X., Shi, C., Hu, B., Song, G., Ye, Y.: Heterogeneous graph
  structure learning for graph neural networks. In: AAAI (2021)

\bibitem{zheng2020heterogeneous}
Zheng, J., Cai, F., Ling, Y., Chen, H.: Heterogeneous graph neural networks to
  predict what happen next. In: COLING (2020)

\bibitem{zheng2021heterogeneous}
Zheng, Y., Lee, V., Wu, Z., Pan, S.: Heterogeneous graph attention network for
  small and medium-sized enterprises bankruptcy prediction. In: PAKDD. pp.
  140--151 (2021)

\end{thebibliography}

\end{document}